\documentclass[10pt,twocolumn,letterpaper]{article}

\usepackage[pagenumbers]{cvpr}  

\usepackage{graphicx}
\usepackage{capt-of}
\usepackage{xcolor}
\usepackage{amsmath,amssymb,amsfonts,dsfont,pifont,bm,bbm,mathrsfs,mathtools,nicefrac}
\usepackage{algorithm,algpseudocode,listings}
\usepackage{booktabs,multirow,adjustbox,diagbox,threeparttable}
\definecolor{citeblue}{RGB}{48,111,186}
\definecolor{azure}{rgb}{0.0, 0.5, 1.0}
\usepackage[pagebackref,breaklinks,colorlinks,citecolor=citeblue,bookmarks=false]{hyperref}
\usepackage[capitalize]{cleveref}
\crefname{section}{Sec.}{Secs.}
\Crefname{section}{Section}{Sections}
\crefname{table}{Tab.}{Tabs.}
\Crefname{table}{Table}{Tables}
\crefname{figure}{Fig.}{Figs.}
\Crefname{figure}{Figure}{Figures}
\crefname{equation}{Eq.}{Eqs.}
\Crefname{equation}{Equation}{Equations}
\def\ie{\emph{i.e.}}

\newcommand{\E}{\mathbb{E}}
\newcommand{\z}{{\rm\bf z}}
\newcommand{\Z}{\mathcal{Z}}
\newcommand{\x}{{\rm\bf x}}
\newcommand{\X}{\mathcal{X}}

\newcommand{\f}{{\rm\bf f}}
\newcommand{\w}{{\rm\bf w}}
\renewcommand{\L}{{\mathcal{L}}}

\newcommand{\ours}{GLeaD~\xspace{}}
\newcommand{\method}{GLeaD\xspace}
\newcommand{\supp}{\textit{Supplementary Material}\xspace}

\newcommand\nonumfootnote[1]{%
\begingroup%
    \renewcommand\thefootnote{}\footnote{\hspace{-3.7pt}#1}%
    \addtocounter{footnote}{-1}%
\endgroup%
}

\begin{document}

\title{GLeaD: Improving GANs with A Generator-Leading Task}

\author{
    Qingyan Bai$^{1*}$\quad
    Ceyuan Yang$^{2}$\quad
    Yinghao Xu$^{3*}$\quad
    Xihui Liu$^4$\quad
    Yujiu Yang$^{1\dagger}$\quad
    Yujun Shen$^5$ \\
    \\
    $^{1}$Tsinghua Shenzhen International Graduate School, Tsinghua University \quad
    $^{2}$Shanghai AI Laboratory \\
    $^{3}$The Chinese University of Hong Kong \quad
    $^{4}$The University of Hong Kong \quad
    $^{5}$Ant Group\\
}

\maketitle

\begin{abstract}

Generative adversarial network (GAN) is formulated as a two-player game between a generator (G) and a discriminator (D), where D is asked to differentiate whether an image comes from real data or is produced by G.
Under such a formulation, D plays as the rule maker and hence tends to dominate the competition.
Towards a fairer game in GANs, we propose a new paradigm for adversarial training, which \textbf{makes G assign a task to D} as well.
Specifically, given an image, we expect D to extract representative features that can be adequately decoded by G to reconstruct the input.
That way, instead of learning freely, D is urged to align with the view of G for domain classification.
Experimental results on various datasets demonstrate the substantial superiority of our approach over the baselines.
For instance, we improve the FID of StyleGAN2 from 4.30 to 2.55 on LSUN Bedroom and from 4.04 to 2.82 on LSUN Church.
We believe that the pioneering attempt present in this work could inspire the community with better designed generator-leading tasks for GAN improvement. Project page is at \url{https://ezioby.github.io/glead/}.

\end{abstract}

\nonumfootnote{$*$ This work was done during an internship at Ant Group.\\
\indent {\hspace{1mm}}$\dagger$ Corresponding author.
This work was partly supported by the National Natural Science Foundation of China (Grant No. U1903213) and the Shenzhen Science and Technology Program (JCYJ20220818101014030).}

\section{Introduction}\label{sec:intro}
Generative adversarial networks (GANs)~\cite{goodfellow2014gan} have significantly advanced image synthesis, which is typically formulated as a two-player game.
The generator (G) aims at synthesizing realistic data to fool the discriminator (D), while D pours attention on distinguishing the synthesized samples from the real ones. 
Ideally, it would come to an optimal solution where G can recover the real data distribution, and D can hardly tell the source of images anymore~\cite{goodfellow2014gan}.

However, the competition between G and D seems to be unfair. 
Specifically, on the one hand, D acts as a \textit{player} in this adversarial game by measuring the discrepancy between the real and synthesized samples.
But on the other hand, the learning signals (\ie,  gradients) of G are only derived from D, making the latter naturally become a \textit{referee} in the competition. 
Such a formulation easily allows D to rule the game. 
Massive experimental results could serve as supporting evidence for the theoretical analysis. 
For instance, in practice, D can successfully distinguish real and fake samples from a pretty early stage of training and is able to maintain its advantage in the entire training process~\cite{wang2021eqgan}. 
Accordingly, the capability of the discriminator usually determines the generation performance more or less. For instance, a discriminator that has over-fitted the whole training set always results in synthesis with limited diversity and poor visual quality~\cite{karras2020ada}.  
Following this philosophy, many attempts~\cite{jolicoeur2018relativistic, yang2021insgen, jeong2021training, shi2022improving, lee2022ggdr, kumari2022ensembling} have been made for discriminator improvement.

\begin{figure}[t]
\centering
\includegraphics[width=\linewidth]{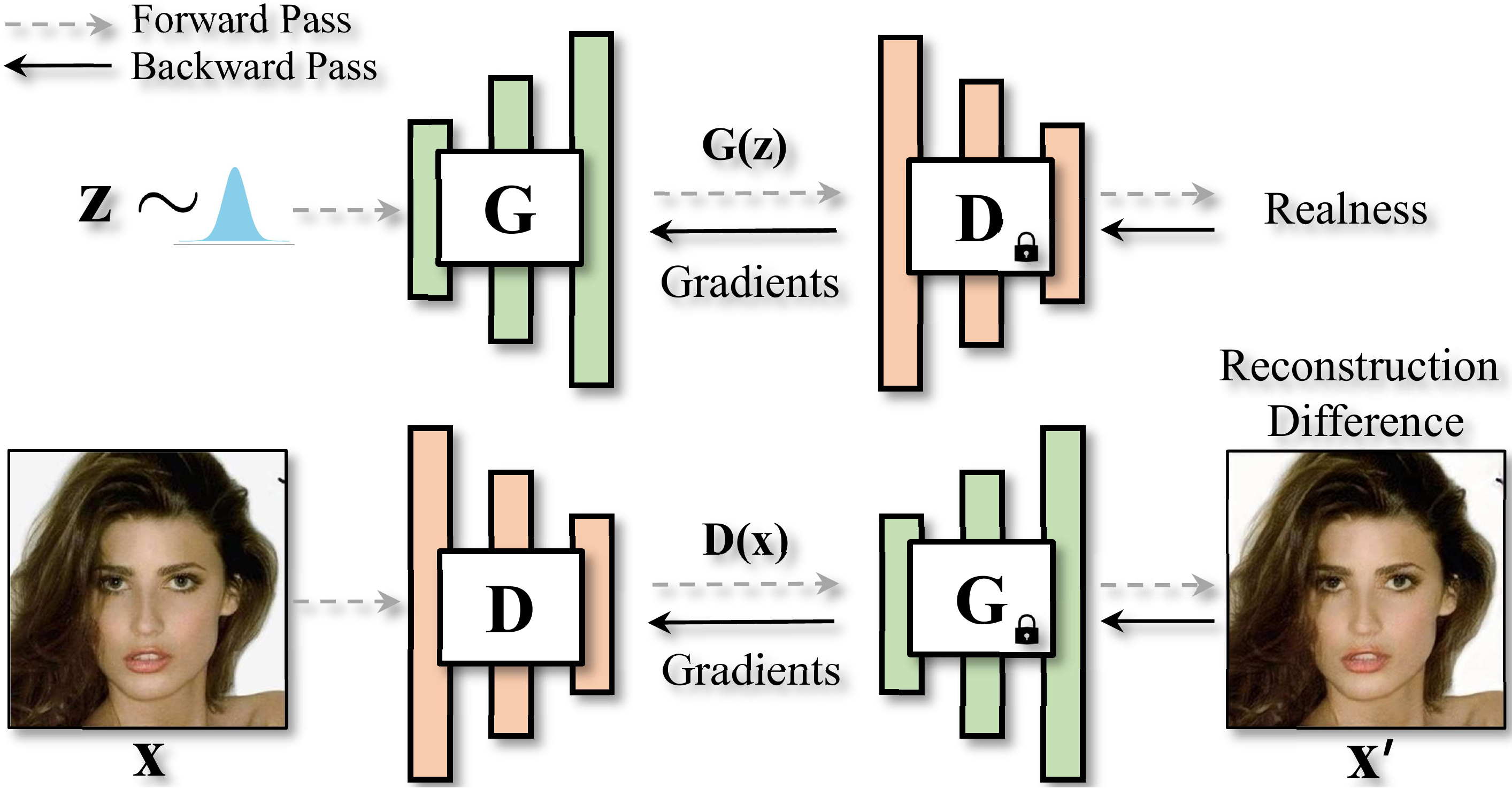}
\caption{
\textbf{Concept diagram} of our proposed generator-leading task (bottom),
as complementary to the discriminator-leading task in the original formulation of GANs (upper).
D is required to extract representative features that can be adequately decoded by G to reconstruct the input.
}
\label{fig:teaser}
\vspace{-5pt}
\end{figure}

This work offers a different perspective on GAN improvement.
In particular, we propose a new adversarial paradigm where G is assigned a new role, \emph{i.e.}, playing as the referee as well to guide D.
Recall that producing realistic images usually requires $G$ to generate \textit{all-level} concepts adequately.
Nevertheless, due to the asymmetrical status of G and D, D is able to tell apart the real and synthesized data merely from limited discriminative regions~\cite{wang2021eqgan}.  
We, therefore, would like to encourage D to extract as much information from an image as possible, such that the features learned by D could be rendered back to the input with a frozen G, as in~\cref{fig:teaser}.
That is, D is enforced to align with the view of G (\textit{i.e.} focusing on the entire image region) instead of learning freely for domain classification.

Our method is termed as \textbf{GLeaD} because we propose to assign D a generator-leading task.
In particular, given a real or synthesized image, the discriminator would deliver extra spatial representations and latent representations that are then fed into a frozen generator to reproduce the original image. 
Reconstruction loss (perceptual loss is adopted in practice) penalties the difference between the input image and the reconstructed image and derives gradients from updating the parameters of the discriminator.
Moreover, comprehensive experiments are then conducted on various datasets, demonstrating the effectiveness of the proposed method. 
Particularly, our method improves Frechet Inception Distance (FID)~\cite{fid} from 4.30 to 2.55 on LSUN Bedroom and 4.04 to 2.82 on LSUN Church. 
We also manage to improve Recall~\cite{kynkaanniemi2019improvedpr} largely (56\%) from 0.25 to 0.39 on LSUN Bedroom.
In addition, thorough ablation studies also suggest that applying generator-leading tasks to require D to reconstruct only real or fake images could boost synthesis quality.
While a larger improvement would be gained if both real and synthesized images were incorporated. 
Last but not least, experimental results in Sec.~\ref{sec:exp} reveal that our method can indeed boost the fairness between G and D as well as improve the spatial attention of D. 
\section{Related Work}\label{sec:related-work}
\noindent\textbf{Generative adversarial networks}.
As one of the popular paradigms for generative models, generative adversarial networks (GANs)~\cite{goodfellow2014gan} have significantly advanced image synthesis~\cite{mirza2014conditional, 
radford2015unsupervised, gulrajani2017improved, brock2018biggan, karras2017progressive, karras2019stylegan, karras2020stylegan2, karras2020training, liu2021FastGAN, karras2021alias, esser2021taming}, as well as various tasks like image manipulation~\cite{shen2020interfacegan, harkonen2020ganspace, tov2021designing, patashnik2021styleclip, wu2021stylespace, zhu2021low}, image translation~\cite{isola2017image, liu2017unsupervised, zhu2017cyclegan, tang2020xinggan, choi2020stargan, wan2020bringing}, image restoration~\cite{agustsson2019generative, gu2020image, pan2021exploiting, wang2021towards}, 3D-aware image synthsis~\cite{zhou2021cips, chan2021pi, gu2021stylenerf, xu2022volumeGAN, shi20223daware}, and talking head generation~\cite{wu2018reenactgan, hong2022depth, yin2022styleheat}.
In the traditional setting of GAN training, D serves as the referee of synthesis quality and thus tends to dominate the competition. 
As a result, in practice D can always tell the real and fake samples apart and the equilibrium between G and D turns out hard to be achieved as expected~\cite{goodfellow2014gan, arjovsky2017wgan}.
Some earlier work~\cite{arjovsky2017wgan, berthelot2017began, fedus2017many} tries to boost GAN equilibrium to stabilize GAN training and improve synthesis quality.
Recently, EqGAN-SA~\cite{wang2021eqgan} proposes to boost GAN equilibrium by raising the spatial awareness of G. 
Concretely, the spatial attention of D is utilized to supervise and strengthen G.
While our method forces D to fulfill a reconstruction task provided by G without improving the capacity of G for the first time. 
To learn useful feature representations with weak supervision, BiGAN~\cite{donahue2016bigan} proposes to learn an encoder, to project real samples back into GAN latent space in addition to the original G and D.
And D is required to discriminate samples jointly in data and latent space.
In this way, the well-trained encoder could serve as a feature extractor in a weak-supervised training manner.
Differently, we directly adopt D to extract features of both real and synthesized samples to reconstruct them with G for a fairer setting instead of representation learning.

\noindent\textbf{Improving GANs with the enhanced discriminator.}
Considering D largely dominates the competition with G, many prior works attempt to boost synthesis quality by improving D.
Jolicoeur employs a relativistic discriminator~\cite{jolicoeur2018relativistic} to estimate the probability that the given real data is more realistic than fake data for better training stability and synthesis quality.
Yang et al.~\cite{yang2021insgen} propose to improve D representation by additionally requiring D to distinguish every individual real and fake image.
Kumari et al.~\cite{kumari2022ensembling} propose to ensemble selected backbones pre-trained on visual understanding tasks in addition to the original D as a strengthened D.  
The effect of various capacity of discriminator on training a generator is also investigated in~\cite{yang2022improving}.
Based on the finding of OASIS~\cite{sushko2020oasis} that dense supervision such as segmentation labels could improve the representation of D in conditional synthesis, 
GGDR~\cite{lee2022ggdr} leverages the feature map of G to supervise the output features of D for unconditional synthesis.
However, different from the discrimination process, G does not backward any gradient to D in this work. 
Contrasted with GGDR, our method aims at a fairer setting rather than gaining more supervision for D.
Also, our D receives gradients from G, leading to fairer competition.

\noindent\textbf{Image reconstruction with GANs.}
GAN inversion~\cite{xia2022gan} aims to reconstruct the input image with a \textit{pre-trained} GAN generator.
Mainstream GAN inversion methods include predicting desirable latent codes corresponding to the images through learning an encoder~\cite{pidhorskyi2020alae, zhu2020idinvert, richardson2021pSp, tov2021e4e, alaluf2021restyle} or optimization~\cite{image2stylegan, abdal2020image2stylegan++, roich2021pivotal, creswell2018inverting, gu2020image, pan2020dgp}.
Most work chooses to predict latent codes in the native latent space of StyleGAN such as $\mathcal{Z}$, $\mathcal{W}$ or $\mathcal{W+}$.
Recently there are also some work~\cite{bai2022high, kang2021bdinvert} extending the latent space or fine-tuing~\cite{dinh2022hyperinverter, alaluf2022hyperstyle} the generator for better reconstruction. 
Note that although our method could achieve image reconstruction with the well-trained D and G, our motivation lies in boosting generative quality by making G assign the generator-leading reconstruction task to D, instead of the reconstruction performance.
Another significant difference lies in that we adopt D to extract representative features for reconstruction, which is simultaneously trained with G, while in GAN inversion the feature extractor (namely the encoder) is learned based on a pre-trained G.

\begin{figure*}[t]
    \centering
    \includegraphics[width=0.98\textwidth]{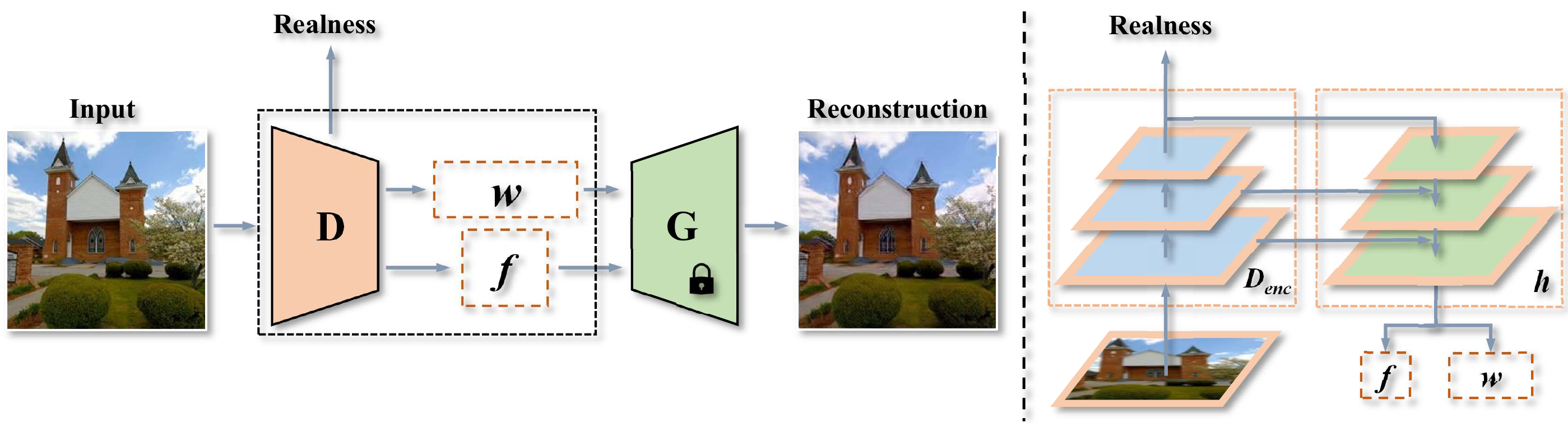}
    \caption{%
    Illustration of how a \textbf{generator-leading task} is incorporated into GAN training from the perspective of discriminator optimization.
    Given an image (\textit{i.e.}, either real or synthesized) as the input, D is asked to extract representative features from the input in addition to predicting a realness score.
    These features including spatial features $f$ and global latent codes $w$ are sent to the fixed G to reconstruct the inputs of D.
    The perceptual loss is adopted to penalize the difference between the reconstruction and inputs.
    The sub-figure on the right demonstrates the specific architecture of our D.
    A decoder $h$ composed of a series of $1\times1$ convolution layers is attached to the original backbone $D_{enc}$ to extract $f$ and $w$. 
    This training process is described in detail in~\cref{subsec:our_task}.
    }
    \label{fig:framework}
    \vspace{-1pt}
\end{figure*}
\section{Method}\label{sec:method}

As mentioned before, it seems to be unfair that a discriminator (D) competes against a generator (G) since D does not only join the two-player game as a \textit{player} but also guides the learning of G, namely serves as a \textit{referee} for G. \cref{subsec:pre} presents the vanilla formulation. 
To chase a fairer game, \cref{subsec:our_task} introduces a new adversarial paradigm \ours that assigns a new generator-leading task for D which in turn is judged by G.

\subsection{Preliminary}\label{subsec:pre}
GAN usually consists of two components: a generator $G(\cdot)$ and a discriminator $D(\cdot)$. The former aims at mapping a random latent code $\z$ to an image, while the latter learns to distinguish the synthesized image $G(\z)$ from the real one $\x$. 
These two networks compete with each other and are jointly optimized with the learning objectives as follows: 
\begin{align}
    \L_G & = - \E_{\z\in\Z}[\log(D(G(\z)))], \label{eq:loss_G} \\
    \L_D & = - \E_{\x\in\X}[\log(D(\x))] - \E_{\z\in\Z}[\log(1-D(G(\z)))],
    \label{eq:loss_D}
\end{align}
where $\Z$ and $\X$ denote a pre-defined latent distribution and data distribution respectively. 

Ideally, the optimal solution is that G manages to reproduce the realistic data distribution while D is not able to tell the real and synthesized samples apart~\cite{goodfellow2014gan}. However, during the iterative training of the generator and discriminator, there exists an unfair competition since D plays the \textit{player} and \textit{referee} roles simultaneously. Thus the ideal solution is hard to be achieved in practice~\cite{wang2021eqgan, farnia2020gans}.

\subsection{Generator-leading Task}\label{subsec:our_task}
Considering the unfair division of labor in this two-player game, we turn to assign a new role to G that could supervise the learning of D in turn.
Recall that the target of generation is to produce realistic samples which usually requires all concepts well-synthesized. 
However, it is suggested~\cite{wang2021eqgan} that the most discriminative regions of given real or synthesized images are sufficient for domain classification. 
Therefore, we propose a generator-leading task that enforces D to extract as many representative features as possible to retain adequate information that could reconstruct a given image through a frozen generator, as described in~\cref{fig:framework} and \cref{tab:algorithm}.
Note that we empirically validate that requiring D to extract spatial  representations is essential to improve synthesis quality in~\cref{subsec:ablation}.
Taking StyleGAN2~\cite{karras2020stylegan2} as an example, we will introduce the detailed instantiations in the following context.

\noindent \textbf{Extracting representations through D.}
The original D of StyleGAN is a convolutional network composed of a series of downsampling convolution layers.
To make it convenient, the backbone network of the original D (namely, parts of D except the final head predicting realness score) is denoted as $D_{enc}$ in the following statement. 
In order to predict the representative features of a given image while retaining various information from low-level to high-level, we additionally affiliate $D_{enc}$ with a decoder $h(\cdot)$ to construct our new D with a multi-level feature pyramid~\cite{lin2017fpn}.
Based on such feature hierarchy ending with a convolutional head, spatial representations $\f$ and latent representations $\w$ are predicted respectively. 
In particular, the newly-attached parts over the backbone adopt convolution layers with the kernel size of $1 \times 1$. 
This is because the crucial part in D that influences the synthesis quality of G is the backbone while introducing too many parameters for $h$ will encourage the optimization to focus on this reconstruction branch (decoder).
Moreover, considering the residual architecture of G, the spatial representation $\f$ consists of a low-level feature and a high-level one in total. More details are available in \supp. Therefore, given one real image $\x$ or a synthesized one $G(\z)$, the corresponding representative features could be obtained by:
\begin{align}
\f_{real}, \w_{real} &= h(D_{enc}(\x)), 
\label{eq:gen_real_fw}\\
\f_{fake}, \w_{fake} &= h(D_{enc}(G(\z)))
\label{eq:gen_fake_fw}.
\end{align}

\noindent \textbf{Reconstructing images via a frozen G.} 
For a fair comparison, the generator of the original StyleGAN2 is adopted without any modification, which stacks a series of convolutional ``synthesis blocks''. 
Notably, the StyleGAN2 generator is designed with a residual architecture, which synthesizes images progressively from a lower resolution to the higher one.
For instance, the $16\times16$ synthesized result of the synthesis block corresponding to a lower resolution is firstly upsampled to $32\times32$, and then the $32\times32$ synthesis block only predicts the residual between the upsampled result and the desirable $32\times32$ image.
As mentioned before, our predicted spatial representations indeed contain two features that could serve as the basis and the residual respectively. 
And the latent representation is sent to the synthesis blocks to modulate the features to generate the final output just as in~\cite{karras2019stylegan, karras2020stylegan2}.
Such that, the reconstructed images could be derived from:
\begin{align}
\x_{real}^\prime & = G(\f_{real}, \w_{real}),  
\label{eq:rec_real}\\
\x_{fake}^\prime & = G(\f_{fake}, \w_{fake}),
\label{eq:rec_fake}
\end{align}
where G is fully frozen.

\noindent \textbf{Reconstruction loss.}
\begin{algorithm}[t] 

\small
\caption{\small{GAN training with the proposed generator-leading task.}}\label{tab:algorithm}
\begin{algorithmic}[1]
\renewcommand{\algorithmicrequire}{\textbf{Input:}}
\renewcommand{\algorithmicensure}{\textbf{Output:}}
\algnewcommand\algorithmicinput{\textbf{Hyperparameters:}}
\algnewcommand\Hyperparameters{\item[\algorithmicinput]}
    \Require  $G$ and our $D$ (including $h$) that are initialized with random parameters. Training data $\{\rm\bf x_i \}$.
    \Hyperparameters $T$: maximum number of training iterations. 

    \For{ $t = 1$ to $T$} 
    \State Sample $ \z \sim  \mathbb{P}(\Z)$
    \Comment{Begin training of $G$.}
    \State Update $G$ with~\cref{eq:loss_G}
    \State Sample $ \z \sim  \mathbb{P}(\Z)$ 
    \Comment{Begin training of $D$.}
    \State Reconstruct $G(\z)$ with~\cref{eq:gen_fake_fw} and~\cref{eq:rec_fake}
    \State Sample $\x \sim \{\rm\bf x_i \} $
    \State Reconstruct $\x$ with~\cref{eq:gen_real_fw} and~\cref{eq:rec_real}
    \State Discriminate images by $D(G(\z))$  and $D(\x)$
    \State Update $D$ with~\cref{eq:loss_D},~\cref{eq:whole_rec_loss},  and~\cref{eq:loss_D_new}
    \EndFor
    \Ensure $G$ with best training set FID.
\end{algorithmic}.
\vspace{-5pt}
\end{algorithm}
After gathering the reconstructed real and synthesized images, we could easily penalize the differences between the original images and reconstructed ones. 
Here, perceptual loss~\cite{zhang2018perceptual} $\L_{per}$ is adopted as the loss function:
\begin{align}
\L_{rec} = \lambda_{1}\L_{per}(\x, \x_{real}^\prime) + \lambda_{2}\L_{per}(G(\z), \x_{fake}^\prime),  
\label{eq:whole_rec_loss}
\end{align}
where $\lambda_{1}$ and $\lambda_{2}$ denote the weights for different terms. 
Note that setting one weight as zero is identical to disabling the reconstruction tasks on real/synthesized images, which may deteriorate the synthesis performance to some extent. 
Our final algorithm is summarized as in~\cref{tab:algorithm}.

\noindent \textbf{Full objective.} With the updated D architecture and the generator-leading task, the discriminator and generator are jointly optimized with
\begin{align}
    \L_G^\prime & = \L_G,  
    \label{eq:loss_G_new}\\
    \L_D^\prime & = \L_D + \L_{rec}
    \label{eq:loss_D_new}.
\end{align}

\section{Experiments}\label{sec:exp}
{
\setlength{\tabcolsep}{9.5pt}
\begin{table*}[t]
\begin{center}
\caption{Comparisons on FFHQ~\cite{karras2019stylegan}, LSUN Bedroom and LSUN Church~\cite{yu2015lsun}. Our method improves StyleGAN2~\cite{karras2020stylegan2} in large datasets in terms of FID~\cite{fid} and recall. P and R denote precision and recall~\cite{kynkaanniemi2019improvedpr}. 
Lower FID and higher precision and recall indicate better performance. 
The bold numbers indicate the best metrics for each dataset.
The blue numbers in the brackets indicate the improvements. 
}
\label{tab:comparision_sota}
\small{
\begin{tabular}{r c c c c c c c c c c c c}
\toprule
\multirow{2}{*}{Method} & \multicolumn{3}{c}{FFHQ~\cite{karras2019stylegan}} & \multicolumn{3}{c}{LSUN Bedroom~\cite{yu2015lsun}} & \multicolumn{3}{c}{LSUN Church~\cite{yu2015lsun}} \\
\cmidrule(lr){2-4}
\cmidrule(lr){5-7}
\cmidrule(lr){8-10}
&FID$\downarrow$ &P$\uparrow$ &R$\uparrow$ 
&FID$\downarrow$ &P$\uparrow$ &R$\uparrow$ 
&FID$\downarrow$ &P$\uparrow$ &R$\uparrow$ \\
\midrule
UT~\cite{bond2021unleashing} 
&  6.11  & \textbf{0.73} & 0.48 
&  -     & - & - 
&  4.07  &  \textbf{0.71} & 0.45 \\
Polarity~\cite{humayun2022polarity} 
& -   & - & - 
&  -  & - & - 
&  3.92 &  0.61 & 0.39 \\
\midrule
StyleGAN2~\cite{karras2020stylegan2}
& 3.79 & 0.68 & 0.44
& 4.30 & 0.59 & 0.25
& 4.04 & 0.58 & 0.40 \\
Ours
& 3.24          (\textcolor{azure}{$-$0.55})
&0.69 &0.47
& \textbf{2.55  (\textcolor{azure}{$-$1.75})}
&\textbf{0.62} &\textbf{0.39} 
& 2.82          (\textcolor{azure}{$-$1.22})
& 0.62 &0.43 \\
\midrule
GGDR~\cite{lee2022ggdr}
& 3.25 & 0.66  & \textbf{0.51}
& 3.71 & 0.62  & 0.33
& 2.81 & 0.61  & 0.46\\
Ours*
&\textbf{2.90 (\textcolor{azure}{$-$0.35})}
&0.69 &0.50 
&2.72         (\textcolor{azure}{$-$0.99})
& 0.62 & 0.37 
&\textbf{2.15 (\textcolor{azure}{$-$0.66})} 
& 0.61 &\textbf{0.48}\\
\bottomrule
\end{tabular}
}
\end{center}
\end{table*}
}

\label{subsec:main_results}
\begin{figure*}[t]
\centering
\includegraphics[width=0.85\textwidth]{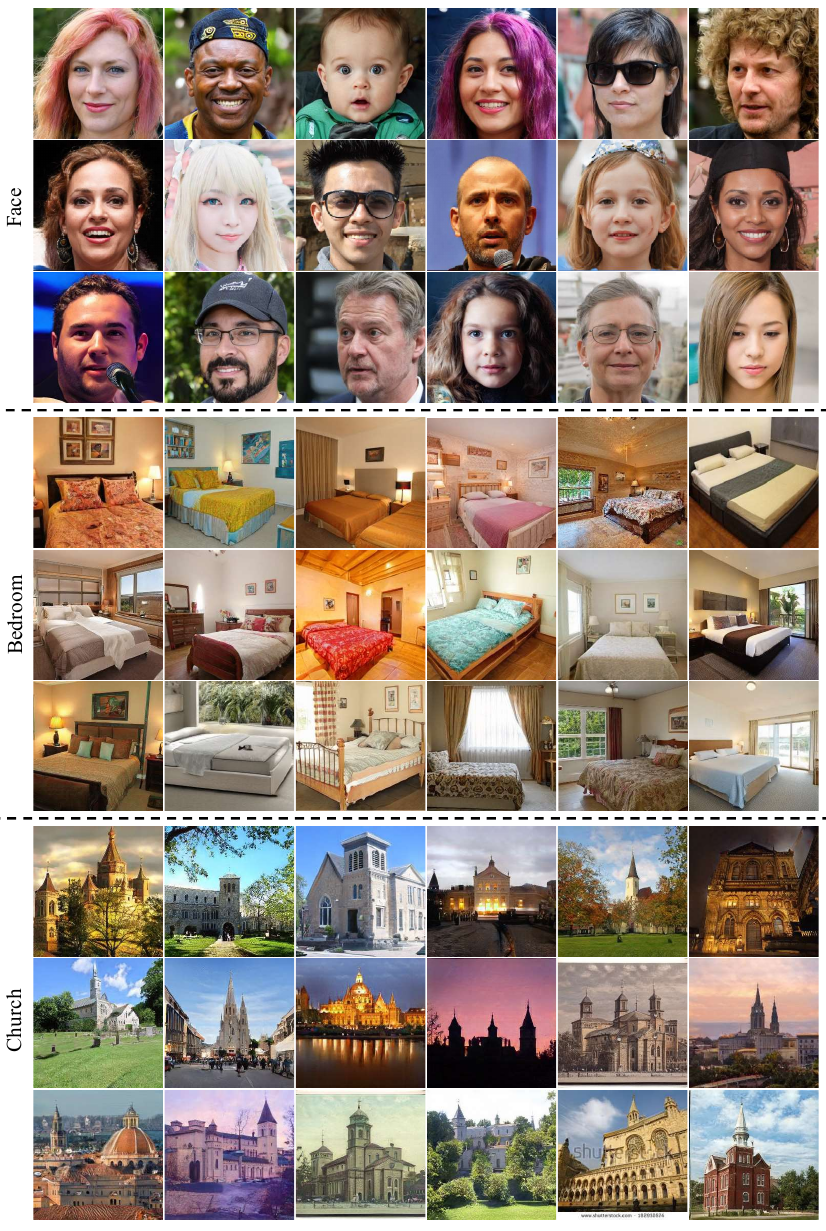}
\vspace{-5pt}
\caption{
Synthesized images by our models respectively trained on FFHQ~\cite{karras2019stylegan}, LSUN Bedroom and Church~\cite{yu2015lsun}.
}
\label{fig:quality}
\vspace{-20pt}
\end{figure*}
We conduct extensive experiments on various benchmark datasets to demonstrate the effectiveness of the proposed method and the superiority of the specific settings.
The subsections are arranged as follows:
\cref{subsec:settings} introduces our detailed experiment settings.
In \cref{subsec:main_results} we demonstrate the qualitative and quantitative superiority of \ours.
\cref{subsec:ablation} includes comprehensive ablation studies of the designed components.
Then we visualize the realness score curves of D to validate the improvement of fairness in~\cref{subsec:eq}.
At last, we provide qualitative reconstruction results and validate the improvement of D's spatial attention respectively in~\cref{subsec:rec_res} and~\cref{subsec:d_attention}.

\subsection{Experimental Setup}
\label{subsec:settings}
\noindent\textbf{Datasets.}
We conduct experiments on FFHQ~\cite{karras2019stylegan} consisting of 70K high-resolution portraits for face synthesis.
We also adopt the training set of LSUN Bedroom and Church~\cite{yu2015lsun} respectively for indoor and outdoor scene synthesis, which respectively contains about 126K and 3M $256\times256$ images.

\noindent\textbf{Evaluation.} We mainly adopt the prevalent  Frechet Inception Distance (FID)~\cite{fid} for evaluation. 
Precision \& Recall (P\&R)~\cite{kynkaanniemi2019improvedpr} is also adopted as the supplement of FID for more grounded evaluation.
In particular, we calculate FID and P\&R between all the real samples and 50K synthesized ones for experiments on FFHQ and LSUN Church. 
While for LSUN Bedroom we calculate FID and P\&R between 50K real samples and 50K synthesized ones because feature extracting of 3M samples is rather costly.

\noindent\textbf{Other settings.}
For all the baseline and our models, on FFHQ we keep training the model until D has been shown 25M images with mirror augmentation.
While models on LSUN Church and Bedroom are trained until 50M images have been shown to D for more sufficient convergence.
We adopt VGG~\cite{simonyan2014vgg} as the
pre-trained feature extractor for perceptual loss calculation.
As for the loss weights, we set $\lambda_{1}$ = 10 and $\lambda_{2}$ = 3.

\subsection{Main Results}
\noindent\textbf{Quantitative comparisons.}
In order to compare our \method against prior works, \emph{e.g.}, UT~\cite{bond2021unleashing}, Polarity~\cite{humayun2022polarity}, and StyleGAN2~\cite{karras2020stylegan2}, we calculate the FID and Precision and Recall~\cite{kynkaanniemi2019improvedpr} (P \& R) to measure the synthesis. In particular, Precision and Recall could reflect the synthesis quality and diversity to some extent. Moreover, considering that recent work GGDR~\cite{lee2022ggdr} also leverages the G to enhance the representations of D, we further incorporate it with our method to check whether exists a consistent gain. 

\cref{tab:comparision_sota} presents the results. From the perspective of FID, our direct baseline StyleGAN2 could be substantially improved with the proposed \method, outperforming other approaches by a clear margin. These results strongly demonstrate the effectiveness of our \method. Moreover, combined with GGDR (Ours* in the table), our \method could further introduce significant gains, achieving new state-of-the-art performance on various datasets. Namely, the proposed \method could be compatible with the recent work GGDR that also considers improving D through G. 

Regarding Precision and Recall,  clear gains are also observed on multiple benchmarks. Importantly, the improvements mainly come from the Recall side, \emph{i.e.}, the synthesis diversity is further improved.
This matches our motivation that the generator-leading task could further urge D to extract more representative features rather than focus on the limited discriminative regions. As a result, G has to synthesize images with a variety of modes to fool D in turn. 
Moreover, the diversity is significantly improved in the LSUN bedrooms from 0.25 to 0.39 (56\%). This may imply that our \method could continuously benefit from the larger-scale reconstruction task, which we leave in future studies.

\noindent\textbf{Qualitative results}.
\cref{fig:quality} presents the synthesized samples by our \method.
The models are respectively trained on FFHQ, LSUN Bedroom, and LSUN Church.
Obviously, all models could generate images with desirable quality and coverage.

\noindent\textbf{Computational costs}.
We evaluate the proposed model in terms of parameter amount and inference time.
The specific results could be found in \supp.

\subsection{Ablation Studies}
\label{subsec:ablation}
\begin{table}[t]
  \caption{Ablation studies on the loss weights $\lambda_{1}$ and $\lambda_{2}$.
  The numbers in bold indicate the best FID in each sub-table.
  }
  \centering
   \begin{subtable}[b]{0.5\linewidth}
     \setlength{\tabcolsep}{8pt}
     \centering
  \begin{tabular}{c c c}
\toprule
$\lambda_{1}$ & $\lambda_{2}$ & FID \\
\midrule
0    & 0 & 4.04 \\
\midrule
100  & 0 & 331  \\
10   & 0 & \textbf{3.10} \\
1    & 0 & 3.27 \\
\bottomrule
\end{tabular}
\end{subtable}\hfill
 \begin{subtable}[b]{0.5\linewidth}
 \setlength{\tabcolsep}{10pt}
 \centering
\begin{tabular}{c c c}
  
\toprule
$\lambda_{1}$ & $\lambda_{2}$ & FID \\
\midrule
0    & 10 & 3.32 \\
\midrule
10   & 10 & 3.15 \\
10   & 3  & \textbf{2.82} \\
10   & 1  & 2.85 \\
\bottomrule
\end{tabular}
\end{subtable}\hfill
  \label{tab:ablation_loss}
\end{table}
\begin{table}[t]
\caption{Ablation studies on the resolution of $\f$.
The upper line indicates the resolution settings and the bottom line concludes the corresponding FID performance.
The number in bold indicates the best FID in the table.
}
\small
\centering
\begin{tabular}{c c c c c c}
\toprule
Baseline & $1\times1$ & $8\times8$ &
$16\times16$ & $32\times32$ & $64\times64$ \\
\midrule
4.04 & 4.68 & 3.27 & 3.01 & \textbf{2.82} & 2.88 \\
\bottomrule
\end{tabular}

\label{tab:ablation_resolution}
\end{table}



\noindent\textbf{Constraint strength}.
Here we ablate the specific target of the proposed generator-leading task on LSUN Church.
Recall that we have $\lambda_{1}$ and $\lambda_{2}$ that respectively control the constraint strength when reconstructing real and fake images in~\cref{eq:whole_rec_loss}.
As shown in the left sub-table of~\cref{tab:ablation_loss}, we first set $\lambda_{1} = \lambda_{2} = 0$ to get the baseline performance.
Then we set $\lambda_{2}$ as 0 and explore a proper $\lambda_{1}$ for only reconstructing real images.
Experiments suggest that an overlarge weight like 100 will make the proposed task interfere with the adversarial training and the model cannot converge.
And 10 turns out to be a proper choice for $\lambda_{1}$, improving FID from 4.04 to 3.10.
The results incorporating the reconstruction of fake images are demonstrated in the sub-table on the right of~\cref{tab:ablation_loss}.
We first set $\lambda_{1} = 0$ and $\lambda_{2} = 10$ to validate that merely reconstructing fake images benefits the synthesis quality.
Then we try to find an appropriate $\lambda_{2}$ when the reconstruction of real images has been incorporated in the task ($\lambda_{1}=10$).
Through the aforementioned experiments, reconstructing both  real and fake images when $\lambda_{1} = 10$ and $\lambda_{2} = 3$ turns out to be the best strategy.

\begin{figure}[t]
\centering
\includegraphics[width=0.46\textwidth]{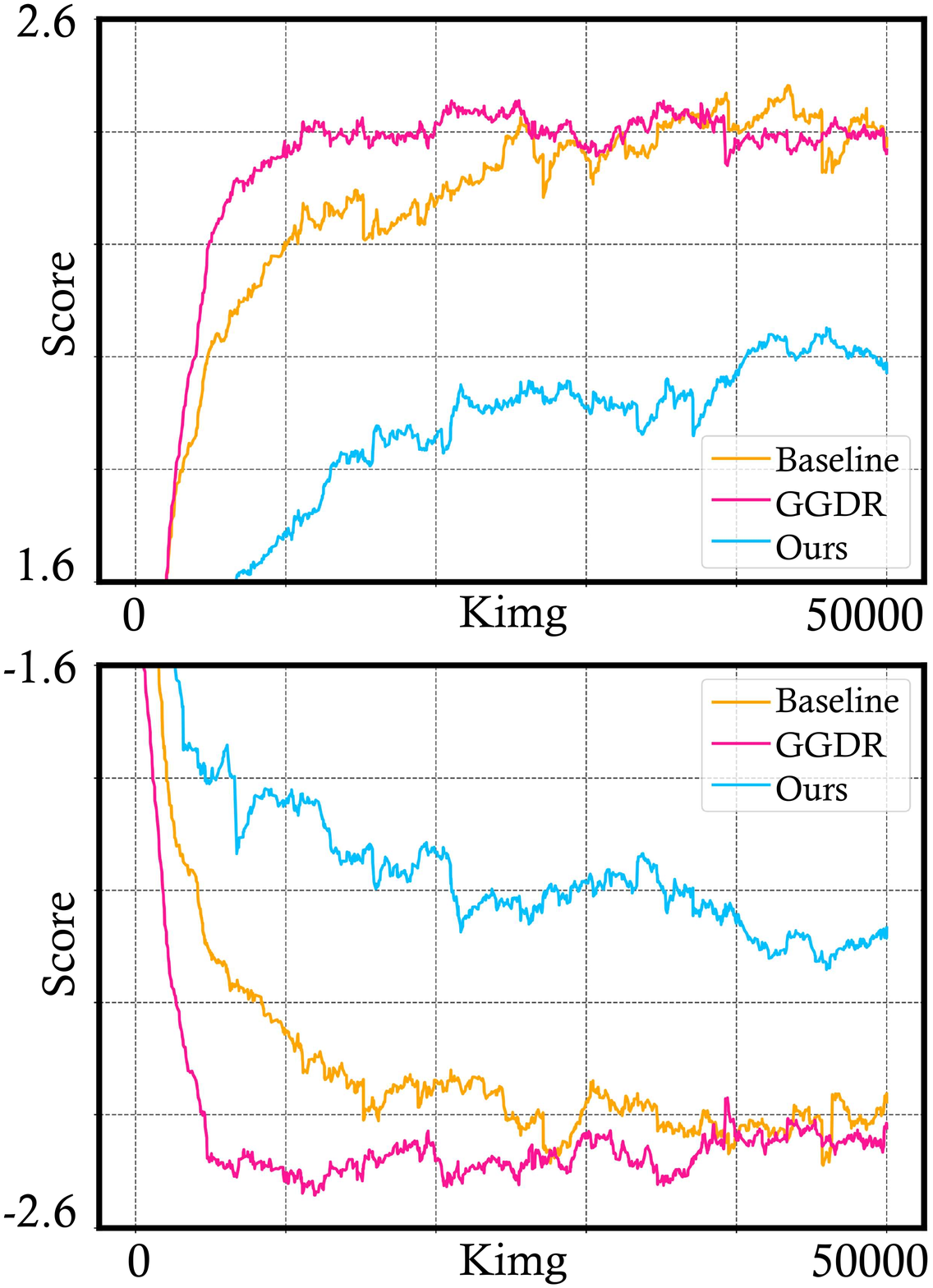}
\caption{
Curves of realness scores that are predicted by various discriminators during training.
The corresponding settings are labeled on the right.
We separately visualize the realness scores from the discriminators of StyleGAN baseline~\cite{karras2020stylegan2}, GGDR~\cite{lee2022ggdr}, and the proposed method.
}
\label{fig:eq}
\end{figure}

\noindent\textbf{Resolution of $\f$}. 
Recall that we require D to extract spatial features $\f$ as the basis of the image reconstruction.
And the predicted latent codes $\w$ modulate the latter features of G to generate the reconstructed image based on $\f$.
Here we conduct ablation studies on the resolution of $\f$ on LSUN Church.
As in~\cref{tab:ablation_resolution}, extracting $\f$ whose resolution is $32\times32$ brings the best synthesis quality.
And $1\times1$ in the table indicates the setting where D only predicts latent codes $w$ without spatial dimension.
Notably, the model performance under this setting is even inferior than the baseline, suggesting the necessity of extracting \textit{spatial} features.

\subsection{Validation of the Fairer Game}
\label{subsec:eq}

\begin{figure}[t]
\centering
\includegraphics[width=0.50\textwidth]{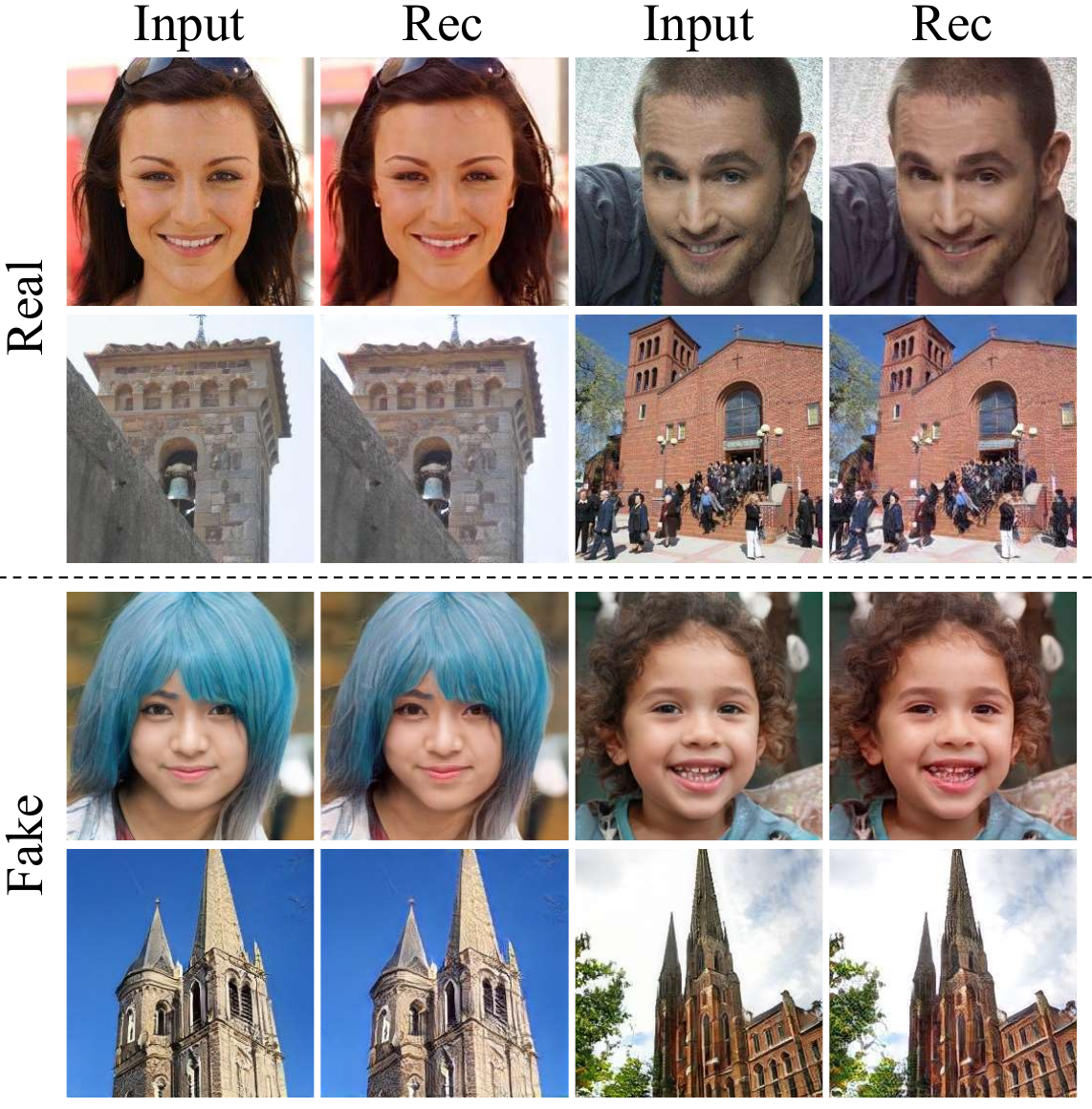}
\caption{
Reconstruction results of real and synthesized input images.
``Input'' and ``Rec'' respectively denote the input images and the reconstruction results by our D and G.
}
\vspace{-10pt}
\label{fig:rec_vis}
\end{figure}
Recall that aiming to improve the synthesis quality through a fairer setting between G and D, we provide the generator-leading task for D to extract representative features adequate for reconstruction. 
Thus we validate the boosted fairness in this subsection through experiments.

Following~\cite{wang2021eqgan}, we visualize the mean score in terms of realness extracted by discriminators throughout the training process on LSUN Bedroom.
Note that the curves are smoothed with exponentially weighted averages~\cite{hunter1986exponentially} for clearer understanding.
The top of \cref{fig:eq} describes the visualization results for the real images while the bottom includes score curves for the synthesized images.
The colors of the curves indicate various settings for training GANs, as labeled on the right of~\cref{fig:eq}.
From the aforementioned figure, it can be found that equipped with our generator-leading task,
the absolute score values of our methods become smaller than the baseline.
While GGDR~\cite{lee2022ggdr} (the red curve) just maintains and even enlarges the gap between the absolute values and zero compared to the baseline, though it can improve FID.

We can thus draw a conclusion that with the aid of the generator-leading task, it becomes much more challenging for D to distinguish the real and fake samples.
In other words, \ours can improve the fairness between G and D, as well as the synthesis quality.
On the contrary, the effectiveness of GGDR is not brought by the improvement of fairness, which emphasizes the viewpoint that, in order to boost fairness between G and D, it is necessary to pass gradients of G to D like our method.

\subsection{Reconstruction Results}
\label{subsec:rec_res}
Recall that we instantiate the generator-leading task as a reconstruction task.
In this subsection, we provide reconstruction results of real and fake images with the well-trained D and G.
To explore the reconstruction ability of D more accurately, we provide it with \textit{unseen} real and synthesized images to extract features.
These features are then fed into the corresponding G to reconstruct the images inputted to D, as in the training stage.
As mentioned in~\cref{subsec:settings}, we train GANs on FFHQ for the face domain and training set of LSUN Church for outdoor scenes.
Thus, here we randomly sample real images from CelebA-HQ~\cite{progan,liu2015faceattributes} (another widely-used face dataset) and the validation set of LSUN Church.
Fake images are sampled with the generators corresponding to the tested discriminators.

As shown in~\cref{fig:rec_vis}, though some out-of-domain objects (\emph{e.g.,} crowds in Church) and high-frequency details (\emph{e.g.,} teeth of the child) are not perfectly well-reconstructed, our well-trained discriminator manages to extract representative features and reproduce the input real and fake images with G.
This indicates that our D could learn features \textit{aligned with the domain of G}, matching our motivation.

\subsection{Spatial Attention Visualization for D}
\label{subsec:d_attention}
\begin{figure}[t]
\centering
\includegraphics[width=0.48\textwidth]{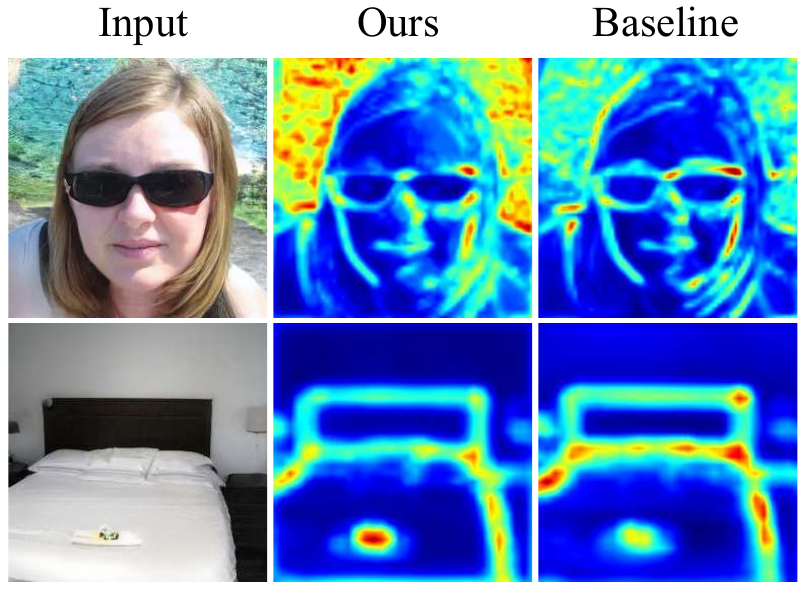}
\caption{
Attention heatmaps of the discriminators visualized by GradCAM~\cite{selvaraju2017grad}.
We feed our D and the baseline D with generated images with artifacts and expect them to pour attention on these regions.
Please zoom in to view the artifacts more clearly. 
}
\label{fig:d_attention}
\end{figure}

We also visualize the spatial attention of the well-trained discriminators with the help of GradCAM~\cite{selvaraju2017grad}.
As mentioned in~\cref{sec:intro}, we expect D to avoid focusing on some limited regions or objects, by extracting spatial representative features. 
Here, the discriminators of the baseline and our method are chosen to validate the improvement in terms of spatial attention.
Considering the discriminators have been fully trained, we pick some generated images with unacceptable artifacts, expecting D aware of these regions with artifacts.
For fair comparison, G of GGDR is adopted to synthesize the images rather than the baseline or ours.
The spatial attention maps are demonstrated in~\cref{fig:d_attention}, note that we pick the gradient map with a relatively higher resolution (64$\times$64) because it is more spatially aligned with the original image than an abstract one (\emph{e.g.,} 8$\times$8).

As in~\cref{fig:d_attention}, the provided fake images contain various kinds of unpleasant artifacts.
The background of the portrait is full of unidentified filamentous  artifacts.
And there is a weird object on the bed in the bedroom picture.
Compared with the baseline D, our D pays much more attention to the artifacts instead of focusing on the face and the bed, which are well synthesized as the subject. 
Recall that under the generator-leading task, D is forced to extract representative spatial features to faithfully reconstruct the inputs.
To achieve this additional task, the backbone of D (namely $D_{enc}$) is naturally forced to learn a much stronger representation than only fulfilling the binary classification task.
Moreover, it is suggested that the strengthened representation of D is strong enough to better detect the generated artifacts.
In contrast, the red regions in the attention map of the baseline are mainly distributed on the face or bed, which means D pays more attention to the subject of the training set, even though there are artifacts generated by G.
Naturally, D's success in detecting and penalizing the artifacts will improve the synthesis capability of G.
And this could serve as one of the reasons why~\ours can boost the synthesis quality of GANs.

\section{Conclusion}\label{sec:conclusion}

Generative adversarial network (GAN) is formulated as a two-player game between a generator (G) and a discriminator (D).
In order to establish a fairer game setting between G and D, we propose a new adversarial paradigm additionally assigning D a generator-leading task, which is termed as GLeaD.
Specifically, we urge D to extract adequate features from the input real and fake images.
These features should be representative enough that G can reconstruct the original inputs with them.
As a result, D is forced to learn a stronger representation aligned with G instead of learning and discriminating freely.
Thus the unfairness between G and D could be alleviated.
Massive experiments demonstrate \ours can significantly improve the synthesis quality over the baseline.

{\small
\bibliographystyle{ieee_fullname}
\bibliography{ref}
}

\appendix
\section*{Appendix}
\newcommand{\blockbegin}[3]{\multirow{5}{*}{\(\left[\begin{array}{c}
\text{1$\times$1 $\mathtt{Conv}$, #1} \\[-.1em]
\text{2$\times$3$\times$3 $\mathtt{Conv}$, #1} \\[-.1em]
\text{1$\times$1 $\mathtt{Conv}$, #1} \\[-.1em] \text{$\mathtt{Downsample}$ } \\[-.1em] \text{$\mathtt{LeakyReLU}$, 0.2} \end{array}\right]\)}
}

\newcommand{\blocks}[3]{\multirow{4}{*}{\(\left[\begin{array}{c}
\text{2$\times$3$\times$3 $\mathtt{Conv}$, #1} \\[-.1em] 
\text{1$\times$1 $\mathtt{Conv}$, #1} \\[-.1em]
\text{$\mathtt{Downsample}$ } \\[-.1em] \text{$\mathtt{LeakyReLU}$, 0.2} \end{array}\right]\)}
}

\newcommand{\blockb}[3]{\multirow{2}{*}{\(\left[\begin{array}{c} 
\text{1$\times$1 $\mathtt{Conv}$, #1} \\[-.1em] \text{$\mathtt{Upsample}$}
\end{array}\right]\)}
}

\newcommand{\blockdec}[3]{\multirow{2}{*}{\(\left[\begin{array}{c} 
\text{1$\times$1
$\mathtt{Conv}$, 3} \\[-.1em]
\text{2$\times$1$\times$1
$\mathtt{Conv}$, #1} \\[-.1em]
\text{$\mathtt{Downsample}$}
\end{array}\right]\)}
}

\newcommand{\blockscore}[3]{
\multirow{7}{*}{\(
\left[\begin{array}{c} 
\text{$\mathtt{Mbstd}$, 1} \\[-.1em] 
\text{3$\times$3
$
\mathtt{Conv}$, #1} \\[-.1em] 
\text{$\mathtt{LeakyReLU}$, 0.2} \\[-.1em]
\text{$\mathtt{Downsample}$} \\[-.1em] 
\text{$\mathtt{FC}$, #1} \\[-.1em] 
\text{$\mathtt{LeakyReLU}$, 0.2} \\[-.1em] 
\text{$\mathtt{FC}$, 1} \\[-.1em]  \end{array}\right]\)}
}

\section{Discriminator Network Structure}\label{sec:supp_arch}

Recall that, our D includes a backbone $D_{enc}$, a head predicting realness scores, and a decoder $h$ for predicting representative features $f$ and $w$.
Taking an image whose resolution is $256\times256$ as an instance, the backbone $D_{enc}$ is first employed to extract features from the input image.
The very last feature map of $4\times4$ is sent to the scoring head to extract the realness score while the multi-level feature maps are sent to the decoder $h$ to predict the representative features adequate for G to reconstruct the original image. 
As described in the submission, the representative features consist of latent codes $w$ and the spatial representations $f$, which include a low-level representation and a high-level representation.
Recall that, these spatial representations will be sent to the fixed generator to serve as the basis of the reconstruction and will be modulated by latent codes to predict the final results.
We illustrate the architectures of the three aforementioned components of D in~\cref{tab:backbone_arch}, \cref{tab:decoder_arch}, and \cref{tab:score_head_arch}, respectively.

\begin{table}[b]
    \centering
    \caption{
        Network structure of the backbone $D_{enc}$.
        The output size is with order $\{C \times H \times W\}$, where $C$, $H$, and $W$ respectively denotes the channel dimension, height and weight of the output.
    }
    \label{tab:backbone_arch}
    \vspace{-8pt}
    \begin{tabular}{ccccccc}
    \toprule
    Stage &  Block &   Output Size  \\ 
    \midrule
    input & - &  $3\times 256 \times 256$ \\
    \midrule
    \multirow{5}{*}{block$_1$} & \blockbegin{{128}}{{64}}{3} & \multirow{5}{*}{$128\times 128 \times 128$} \\
    & &  \\
    & &  \\
    & &  \\
    & &  \\
    \midrule
    \multirow{3}{*}{block$_2$} & \blocks{{256}}{{128}}{4}  &  \multirow{3}{*}{$256\times 64 \times 64 $} \\
    & &  \\
    & &  \\
     & &  \\
    \midrule
    \multirow{4}{*}{block$_3$} & \blocks{{512}}{{256}}{6} &
    \multirow{4}{*}{$512\times 32 \times 32 $} \\
    & &  \\
    & &  \\
     & &  \\
     \midrule
    \multirow{4}{*}{block$_4$} & \blocks{{512}}{{256}}{6} &
    \multirow{4}{*}{$512\times 16 \times 16 $} \\
    & &  \\
    & &  \\
     & &  \\
          \midrule
    \multirow{4}{*}{block$_5$} & \blocks{{512}}{{256}}{6} &
    \multirow{4}{*}{$512\times 8 \times 8 $} \\
    & &  \\
    & &  \\
     & &  \\
               \midrule
    \multirow{4}{*}{block$_6$} & \blocks{{512}}{{256}}{6} &
    \multirow{4}{*}{$512\times 4 \times 4 $} \\
    & &  \\
    & &  \\
     & &  \\
    \bottomrule
  \end{tabular}
\end{table}

\begin{table}[t]
    \centering
    \caption{
        Network structure of the decoder $h$ predicting the low-level spatial representation, the high-level spatial representation and the 512-channel latent codes.
        Note that $h$ receives multi-level features as inputs due to its feature pyramid architecture~\cite{lin2017fpn}.
        The output size is with order $\{C \times H \times W\}$.
    }
    \label{tab:decoder_arch}
    \vspace{-8pt}
    \begin{tabular}{ccccccc}
    \toprule
    Stage &  Block &   Output Size  \\ 
    \midrule
    \multirow{4}{*}{input} & 
    \multirow{4}{*}{$-$} & $512\times32\times32$ \\
    & & $512\times16\times16$ \\
    & & $512\times8\times8$ \\
    & & $512\times4\times4$ \\
    \midrule
    \multirow{2}{*}{block$_1$} & \blockb{{512}}{{64}}{3} & \multirow{2}{*}{$512\times8\times8$} 
    \\
    & &  \\
    \midrule
    \multirow{2}{*}{block$_2$} & \blockb{{512}}{{128}}{4}  &  \multirow{2}{*}{$512\times16\times16$} \\
    & &  \\
    \midrule
    \multirow{2}{*}{block$_3$} & \blockb{{512}}{{256}}{6} &
    \multirow{2}{*}{$512\times32\times32$} \\
    & &  \\
    \midrule
    \multirow{3}{*}{block$_4$} & \blockdec{{512}}{{256}}{6} & $3\times32\times32$ \\
    & & $512\times32\times32$ \\
    & & $512$ \\
    \bottomrule
  \end{tabular}
\end{table}

\begin{table}[t]
    \centering
    \caption{
        Network structure of the head predicting realness scores which are scalars.
        The output size is with order $\{C \times H \times W\}$.
    }
    \label{tab:score_head_arch}
    \vspace{-8pt}
    \setlength{\tabcolsep}{9pt}
    \begin{tabular}{ccccccc}
    \toprule
    Stage &  Block &   Output Size  \\ 
    \midrule
    input & $-$ &  $512\times 4 \times 4$ \\
    \midrule
    \multirow{7}{*}{block$_1$} & \blockscore{{512}}{{64}}{3} & \multirow{7}{*}{1} \\
    & &  \\
    & &  \\
    & &  \\
    & &  \\
    & &  \\
    & &  \\
    \bottomrule
  \end{tabular}
\end{table}

\begin{table}[!ht]
\small
\caption{Computational cost comparisons.}
\label{tab:costs}
\vspace{-8pt}
\centering
{
\begin{tabular}{c c c c}
\toprule
Method & \# params &  inference time(s) & training time(h)\\ 
\midrule
Baseline & 24.00M & 0.0184 & 43.83\\
GLeaD & 25.77M & 0.0219 & 55.78\\
\bottomrule
\end{tabular}
}
\end{table}

\section{Computational Costs}
\label{sec:supp_costs}

We first compute the discriminator parameter amounts of the baseline and our method. 
As in~\cref{tab:costs}, our method merely brings 7.4\% additional parameters over baseline,
which is brought by the proposed lightweight design of $h$ composed of $1\times1$ convolutions.
Then we compare the inference time of the discriminators with a single A6000 GPU.
At last, we make comparisons on the training time.
We separately train the baseline model~\cite{karras2020stylegan2} and our model with 8 A100 GPUs on LSUN Church and record how much time the training costs.
From the numbers in~\cref{tab:costs}, we can conclude that our method improves the synthesis quality without much additional computational burden.

\end{document}